# Artificial Intelligence Methods Based Hierarchical Classification of Frontotemporal Dementia to Improve Diagnostic Predictability

Km Poonam[1, 4], Rajlakshmi Guha[2, 4], Partha P Chakrabarti[3, 4]

[1]Centre of Excellence in Artificial Intelligence, poonamk@iitkgp.ac.in, [2]Rekhi Centre of Excellence for the Science of Happiness, rajg@cet.iitkgp.ac.in, [3]Department of Computer Science & Engineering, and Jointly with the Centre of Excellence in Artificial Intelligence, ppchak@cse.iitkgp.ac.in , [4]IIT Kharagpur, West Bengal, India

*Abstract-* **Patients with Frontotemporal Dementia (FTD) have impaired cognitive abilities, executive and behavioral traits, loss of language ability, and decreased memory capabilities. Based on the distinct patterns of cortical atrophy and symptoms, the FTD spectrum primarily includes three variants: behavioral variant FTD (bvFTD), non-fluent variant primary progressive aphasia (nfvPPA), and semantic variant primary progressive aphasia (svPPA). The purpose of this study is to classify MRI images of every single subject into one of the spectrums of the FTD in a hierarchical order by applying data-driven techniques of Artificial Intelligence (AI) on cortical thickness data. This data is computed by FreeSurfer software. We used the Smallest Univalue Segment Assimilating Nucleus (SUSAN) technique to minimize the noise in cortical thickness data. Specifically, we took 204 subjects from the frontotemporal lobar degeneration neuroimaging initiative (NIFTD) database to validate this approach, and each subject was diagnosed in one of the diagnostic categories (30 bvFTD, 41 svPPA, 25 nfvPPA, 24 Alzheimer's Disease, and 84 cognitively normal). Our proposed automated classification model yielded classification accuracy of 86.5%, 76%, and 72.7% with support vector machine (SVM), linear discriminant analysis (LDA), and Naive Bayes methods, respectively, in 10-fold cross-validation analysis, which is a significant improvement on a traditional single multi-class model with an accuracy of 82.7%, 73.4 %, and 69.2%.**

## I. INTRODUCTION

Frontotemporal dementia is caused due to neuronal degeneration in the frontal and temporal lobes of the brain. These brain lobes control some essential cognitive functions, namely, social cognition, language, emotion, memory, and decision-making abilities [1, 2]. Frontotemporal dementia has primarily three clinical subtypes namely, behavioral variant frontotemporal dementia (bvFTD), non-fluent variant primary progressive aphasia (nfvPPA), and semantic variant primary progressive aphasia (svPPA) [3, 4]. Among them, bvFTD is the most common subtype covering approximately 70% of all FTD patients [3]. Patients with bvFTD manifest changes in behavior and language [5]. Primary progressive aphasia (PPA) is a neurological syndrome in which language capabilities become gradually and progressively impaired [6]. PPA is associated with two language disorders; specifically, nfvPPA, a patient with nfvPPA will have decreased verbal output and, after a certain period, becomes non-verbal. Another disorder is svPPA; this subtype's symptoms are progressive loss of words and semantic knowledge [7, 8].

Magnetic Resonance Imaging (MRI) scans show abnormal neuron degeneration patterns in cognitively impaired patients that can help clinicians diagnose dementia patients. In patients with bvFTD, structural and functional neuroimaging techniques have highlighted cortical atrophy in the orbitofrontal cortex, dorsomedial frontal cortex, and frontoinsular brain regions [9]. The cortical atrophy in nfvPPA patients is correlated with frontal and insular regions [9] and, the similarly for svPPA, it is correlated with anterior temporal lobes, anterior hippocampus, and Amygdala [10].

In medical practice, although a single-subject classification model would be more informative than group analysis. The subtle pattern of cortical atrophy in dementia patients and overlying features of cortical atrophy patterns in the subtypes of FTD patients and other forms of dementia motivate an automated analysis of the image used in the classification of each subject at the separate level. In the study [11], the authors computed the individual network for every subject using the support vector machines (SVM) classifier on cortical thickness data and every subject was classified into one of the diagnostic categories namely, mild cognitive impairment, Alzheimer's disease, and control group. AI-based methods have been popular classifiers for brain images. Many studies have been shown that the SVM classification method using cortical thickness data can differentiate the PPA group from the control group [10, 12]. SVM methods to detect Alzheimer's disease (AD) using surface-based features, specifically, cortical thickness, cortical and subcortical volume, are reported [13]. It has been shown in the study [14] that automated cortical thickness computation from MRI images can distinguish AD patients from the control group. Some studies have explored AI methods to discriminate FTD from cognitively normal (CN) groups using MRI data [15, 16]. MRI-based cortical thickness methods have been proposed in [17] to discriminate the PPA variants using the Random Forest method. So far, classifications of the FTD subtypes, detection of Alzheimer's disease, and discrimination of PPA group have been explored using AI methods. However still, an automated hierarchical classification of the FTD subtypes has not been explored yet.

This study presents a hierarchical classification model that will classify each subject into one diagnostic label (CN, Non-FTD, bvFTD, nfvPPA, and svPPA). The underlying AI model is constructed based on the clinician decision process. If a clinician notices any unusual findings that a normal aging process can't describe in the pre-screening assessment, the clinician must exclude dementia. If the subject has a change in behavior or language that indicates FTD, a clinician typically identifies which disorder it is. We developed an automated classification model to imitate this process. In the proposed method, first cognitively control subjects are separated from dementia patients. Then dementia patients are classified into Non-FTD and FTD groups. Further, the group of FTD patients is classified into behavioral variant and primary progressive aphasia. Eventually, the Primary progressive aphasia group classifies subjects into non-fluent PPA and semantic PPA groups. Using mean cortical thickness, we computed a pair of cortical regions' distances to acquire the subject's connectivity matrix. Then, we applied a kernel function to obtain the connection weight. The proposed hierarchical model uses the Smallest Univalue Segment Assimilating Nucleus (SUSAN) [18] noise reduction technique to smoothen images. Principal component analysis (PCA) was used for dimension reduction and various AI methods (SVM, LDA and Naive Bayes) were used for classification at subsequent levels. Then, we compared the performance of these three methods and the performance of a non-hierarchical classifier.

This study contributes in the following ways; first, our automated hierarchical classification model will classify every patient into one of the diagnostic labels by employing AI methods on surface-based cortical thickness feature only. Second, our proposed method gives better predictability of diagnostic categories than the existing methods (non-hierarchical). Third, the noise removal SUSAN technique has improved the performance of the model. Fourth, our model can assist clinicians in diagnosing FTD subtypes, thus making it easy for them to timely select reasonable interventions.

## II. MATERIALS

### A. Dataset

The dataset used in this study was obtained from Alzheimer's disease neuroimaging initiative (ADNI) study's project, namely, frontotemporal lobar degeneration neuroimaging initiative (NIFTD) [19]. The data access committee approved access to the NIFTD data. Sample images from the dataset have been shown in Figure 1. The dataset contains information of 84 cognitively normal subjects (age ± SD = 68.3±5.51, MMSE ± SD=28.8±1.21 and year of education ± SD = 13.9±3.10), 96 FTD patients (30 bvFTD, 41 svPPA and 25 nfvPPA) (age ± SD = 68.5±6, MMSE±SD = 23.2±6.7 and year of education± SD = 13.4±2.96), and 24 AD patients (age ± SD = 72.1±5, MMSE±SD = 24±6 and year of education ±SD = 13.0±3.1). In the dataset, 99 (48.5%) were men, and 105 (51.4%) were women.

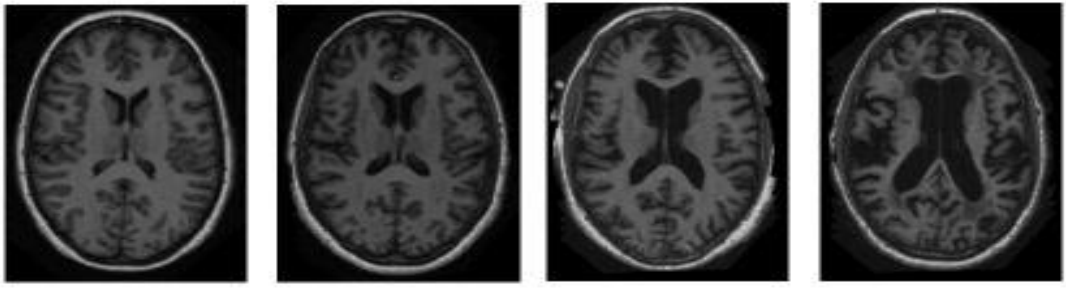

**Fig.1. Sample Images from NIFTD Dataset**

B. Image Acquisition

For every subject, distinct T1-weighted MRI scans were obtained using the 3-Tesla MRI scanner vision in the imaging session and then following features were extracted, viz. TR/TE 8200/86ms, Orientation Sagittal, slice thickness 1 mm and isotropic voxels 2.2 mm$^3$, matrix= 240x256x156 (mm$^3$).

III. Proposed Methodology

The obtained 3D T1-weighted MRI scans need to be processed and analyzed. To do so, we have divided our proposed approach into two parts: First is MRI Image Preprocessing, compute the surface-based feature cortical thickness, and remove the noise in data. The second one is the Hierarchical Classifier which will classify the subjects into different diagnostic categories based on their MRI scans' cortical thickness, computed in the first part. The following two sections A and B, contain a detailed description of both the proposed approach parts. The overall architecture of the same is shown in Figure 2.

A. MRI Image Preprocessing

An open-source software FreeSurfer was used for preprocessing and analyzing MRI images. To compute the features, we followed a certain number of preprocessing steps which are mentioned below. The first step is intensity normalization, which attempts to correct fluctuations in intensity. Subsequently, the second step is skull-stripping, and the Watershed Algorithm [20] was used to remove the skull and anything else from the image that is not grey or white matter, e.g., eyes, neck and, dura mater. Volumetric labeling identified and labeled several subcortical structures such as putamen, hippocampus, ventricles, for attaining correlation among subjects and maintaining the same number of vertices in each hemisphere, the mesh vertices were computed [21, 22]. The fourth step is white matter segmentation which segments white matter from grey matter and cerebrospinal fluid. It also detects if there is any injury, tumor, or other lesions. A detailed explanation of all these preprocessing steps mentioned in the Appendix. Cortical thickness is known to be a good morphological index of the cerebral cortex for tissue classification. Cortical thickness was measured in terms of Euclidian distance between the inner (grey matter/white matter) and outer (pial) cortical surfaces [23]. Cortical thickness computation was measured using methods described in studies [24, 25]. The mean values of cortical thickness data for segmented neuroanatomical regions manifest as a feature vector. For differentiation of the cortical thicknesses of congruous brain regions among the subjects, the thickness was specially registered impartially on the iterative group by matching the sulcal folding pattern using a spherical surface-based registration algorithm [26-28]. The proposed hierarchical classification model uses the SUSAN noise reduction technique with higher smoothing Gaussian kernel full-width at half-maximum (FWHM) of size 15mm for surface-based measurement of the brain to smoothen images. For implementation purposes, we used FreeSurfer's FSL library [29]. Extracted features were fed to the classifier for classification.

B. Hierarchical Classification Using Cortical Thickness Data

For classifying the subjects into different diagnostic categories, we proposed a hierarchical classification approach. This approach uses supervised learning-based AI methods like SVM, LDA, and Naive Bayes [13, 30, and 31]. The proposed model is fully automated, so the input MRI image is fed only once at the hierarchy's top level. As the algorithm proceeds, the required features will be propagated at each subsequent level, and the subject will be classified into more specific clinical labels. Figure 3 shows a detailed architecture of the proposed hierarchical approach. In the first step, our model was trained with the CN and dementia subjects collectively. The Non-FTD group includes the forms of dementia that are not FTD in the dataset, such as AD or vascular dementia, etc. In continuation, the FTD classifier model was trained with behavioral variant and primary progressive aphasia groups. Eventually, the primary progressive aphasia group was trained to differentiate between semantic variant and non-fluent variant. By these four classifiers, the classification was carried out in a hierarchical order. This hierarchical approach was executed iteratively throughout the complete model until each subject was classified into any of the diagnostic categories (CN, Non-FTD, bvFTD, nfvPPA and, svPPA). Gaussian kernel FWHM size of 15mm was applied on cortical thickness data for removal of noise. This approach transforms cortical thickness data's geometrical-domain into frequency-domain [32, 33]. In the process, a high-frequency component is considered as noise and a low-frequency component as a feature. For every subject, the surface of each hemisphere consists of 40,962 vertices and 81,920 mesh triangles. These vertex links were further transformed into 280-dimensional frequency components data. The detailed noise removal technique is described in the studies [34-36]. Finally, the classification model was constructed by employing PCA for dimension reduction; the Supplementary Table_1 contains the information of feature dimension. AI methods, namely, SVM, LDA, and Naïve Bayes, were applied for classification on cortical thickness data, as described in Figure 2. We have applied a noise filtering technique, SUSAN, to enhance the accuracy of our model further. The robustness of the models were estimated by the k-fold cross-validation method. It worked by splitting the dataset into a k=10 subset for each step in the entire hierarchical classification process. Six subsets were utilized for training, two for testing and the other two for validation. To compute the mean accuracy, sensitivity, and specificity, and for the reliability of the model, the cross-validation method was executed 1000 times.

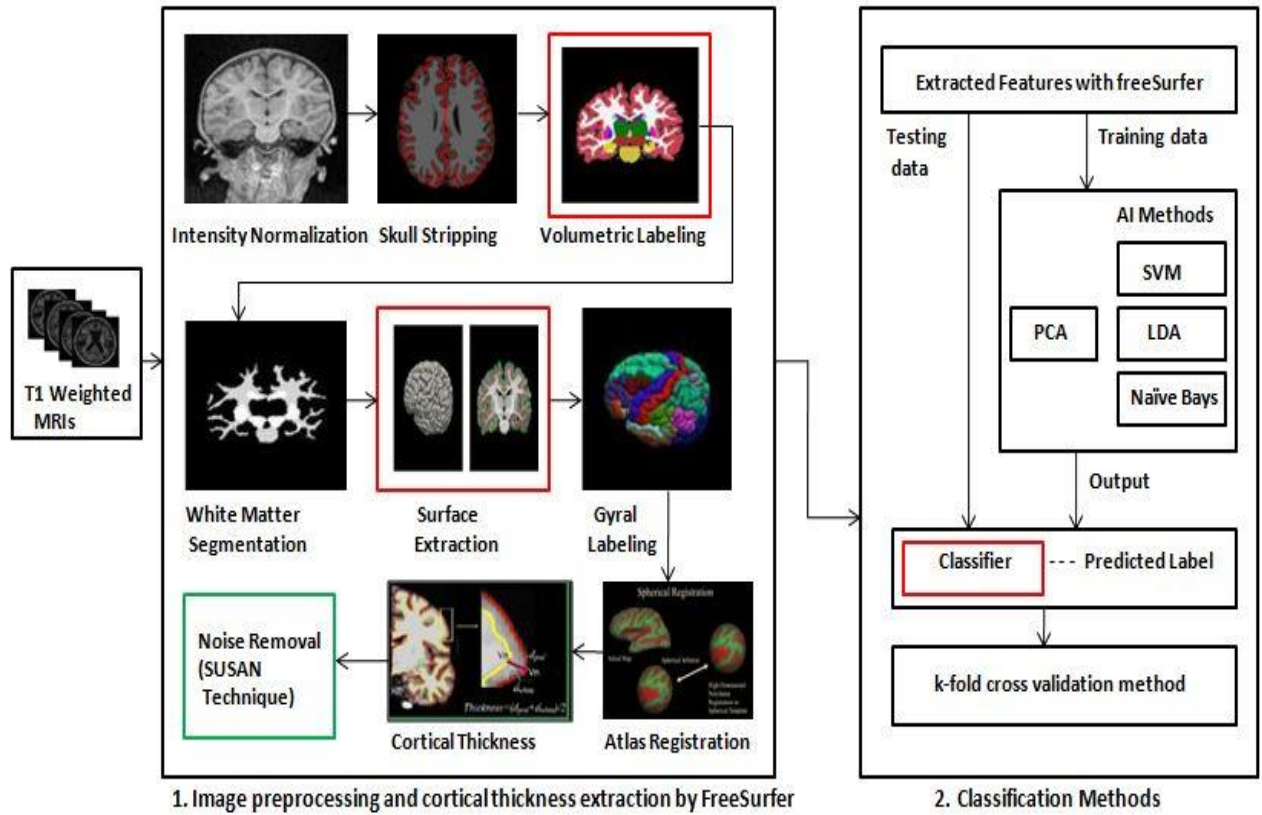

**Fig.2. Proposed System Architecture**

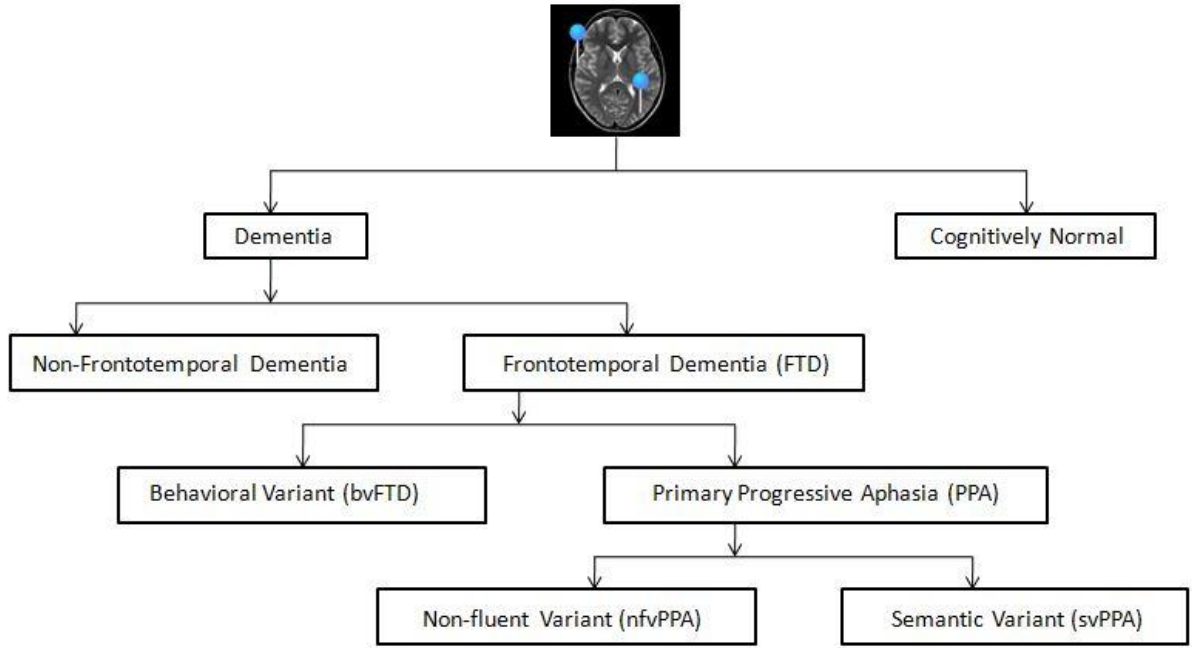

**Fig.3. Hierarchical Classifier**

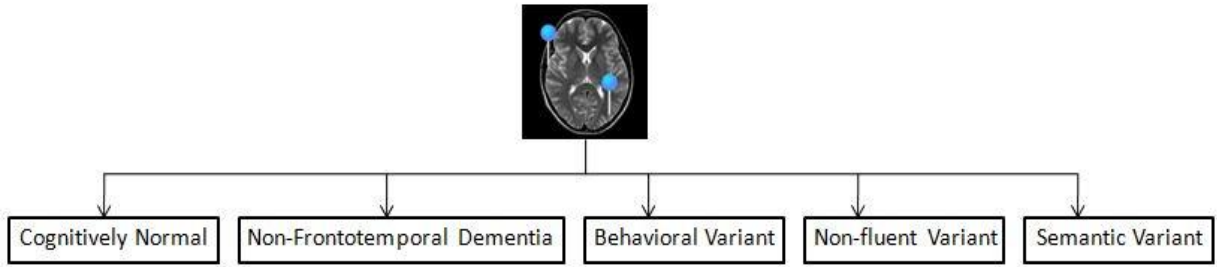

**Fig.4. Single, five-class Classifier**

## IV. RESULTS

### A. Classification Performance

For comparison of classification methods, computed two models. The first is a four-level automated hierarchical classification model as shown in Figure 3, and the second is a single, five-class classification model presented in Figure 4. Table_1 shows the performance measures of each classification step. The automated hierarchical classifier presented in Figure 3 classified every single subject into one of the five clinical groups (CN, Non-FTD, bvFTD, nfvPPA, and svPPA) in the differential manner and obtained a classification accuracy of 86.5%, 76%, and 72.7% with SVM, LDA, and Naïve Bayes methods respectively. A single, multi-class classifier presented in Figure 4 classified each subject into one of the five classes and obtained an accuracy of 82.7%, 73.4%, and 69.2% with SVM, LDA, and Naïve Bayes, respectively. Among all the methods, SVM performs well for both hierarchical and non-hierarchal models. The accuracy of the first step, which discriminates cognitively normal subjects to the demented patients, was 98.5%. In the second step, FTD and Non-FTD patients were classified with an accuracy of 92.3%. In the third step, a group of FTD patients was distinguished between behavioral variant and PPA group with an accuracy of 86.5%. It is critical to discriminate between behavioral variant and primary progressive aphasia groups. Some patients with PPA experience prominent changes in behavior similar to those seen in bvFTD in later stages of the disorder. So, the neuropsychiatric profiles for both the subtypes PPA and bvFTD may converge by last–stage dementia.

| AI Methods | Accuracy | Sensitivity | Specificity |
|---|---|---|---|
| **SVM** | | | |
| Hierarchical tree approach | 86.5% | 88.1% | 80.0% |
| Step1 (CN vs Dementia ) | 98.5% | 100% | 97.6% |
| Step2 (FTD vs Non-FTD) | 92.3% | 95.2% | 80.0% |
| Step3 (bvFTD vs PPA) | 86.5% | 76.9% | 94.5% |
| Step4 (nfvPPA vs svPPA) | 91.8% | 94.4% | 89.5% |
| A single, multi-class approach | 82.7% | 85.7% | 70.0% |
| **LDA** | | | |
| Hierarchical tree approach | 76.0% | 69.3% | 93.1% |
| Step1 (CN vs Dementia) | 88.1% | 91.6% | 86.5% |
| Step2 (FTD vs Non-FTD) | 86.5% | 90.4% | 70.0% |
| Step3 (bvFTD vs PPA) | 84.0% | 76.9% | 86.4% |
| Step4 (nfvPPA vs svPPA) | 89.1% | 88.8% | 89.4% |
| A single, multi-class approach | 73.4% | 67.2% | 92.7% |
| **Naïve Bayes** | | | |
| Hierarchical tree approach | 72.7% | 58.3% | 80.9% |
| Step1 (CN vs Dementia) | 82.8% | 79.1% | 84.6% |
| Step2 (FTD vs Non-FTD) | 84.6% | 88.0% | 70.0% |
| Step3 (bvFTD vs PPA) | 82.0% | 77.0% | 83.7% |
| Step4 (nfvPPA vs svPPA) | 83.7% | 89.0% | 78.9% |
| A single, multi-class approach | 69.2% | 71.4% | 60.0% |

**Table 1: Classification Performance**

Longitudinal data might provide a complete prognosis and evoke risk factors for developing changes over time. In comparison with PPA, the discriminative regions in the bvFTD group primarily showed right frontal. In the fourth step, within the PPA group, nfvPPA and svPPA patients were discriminated against with an accuracy of 91.8%. The overall accuracy of the SVM method's hierarchical classifier is 86.5%, which is a significant improvement on a single, five-class classifier that classifies each subject into one of the clinical groups with an accuracy of 82.7%. The receiver operating characteristics curves (ROC) are plotted for steps 1 to 4 for SVM, LDA, and Naïve Bayes methods. These ROC curves are shown in Supplementary Figures (Fig.1, Fig.2, and Fig.3). The SVM classifier performed better than other methods because of mainly two reasons: First, it finds a decision boundary that maximizes the two groups' margin, and second, the group separating hyperplane, used in LDA, is sensitive to all the data in each class, whereas, in SVM, it is sensitive only to the data on the boundary. Thus, it provides computational savings and makes it more robust to outliers. Python programming language was used to implement the models. Table_2 shows the performance measures of hierarchical and non-hierarchical models using noise removal techniques SUSAN and Laplace-Beltrami operators. SUSAN technique gives additional 1-2% accuracy as compared to the Laplace-Beltrami operator.

| AI Methods | Accuracy with Laplace-Beltrami operator | Accuracy with SUSAN technique |
|---|---|---|
| **SVM** Hierarchical Classifier | 85.3% | 86.5% |
| Non-hierarchical Classifier | 81.2% | 82.7% |
| **LDA** Hierarchical Classifier | 75.1% | 76.0% |
| Non-hierarchical Classifier | 71.9% | 73.4% |
| **Naive Bayes** Hierarchical Classifier | 71.4% | 72.7% |
| Non-hierarchical Classifier | 68.1% | 69.6% |

**Table 2: Performance Measures of Noise Removal Techniques**

B. Misclassification of Subjects

Although our hierarchical classifier classified the subjects with good accuracy. Still some subjects were classified incorrectly in single or several steps; AI methods, namely, SVM, LDA, and Naïve Bayes, try to find out different patterns of cortical atrophy still the same patterns of atrophy among them may guide to misclassification.
For assessing misclassified subjects, neuroanatomists visually reviewed MRI images and found that eight of them had minute cortical atrophy that was not suggestive of any syndrome. Two images manifested specific cortical atrophy, but its spatial pattern was shared by more than one diagnostic category. For some misclassified subjects, there is no proper evidence why misclassification occurred. To handle the misclassification, we computed the automated hierarchical classification model using other hierarchies (step1: CN vs. PPA, step2: non-FTD vs. FTD, step3: dementia vs. svPPA, step4: nfvPPA vs. bvFTD and step1: CN vs. FTD, step2: bvFTDA vs. non-FTD, step3: PPA vs. svPPA, step4: nfvPPA vs. dementia) as well. These models gave poor results than our proposed model hierarchy. One possible way to resolve this kind of overfitting problem in supervised learning is to train the model on a larger dataset. The complete description of the misclassified subjects is mentioned in Supplementary Table_2.

V. CONCLUSION

We proposed an automated hierarchical classification model that classified each subject into one of the diagnostic categories (CN, Non-FTD, bvFTD, svPPA, and nfvPPA) using AI methods on cortical thickness data. SUSAN technique was used for noise removal that improved the performance of the model. The proposed approach has been implemented via applying AI methods, namely, SVM, LDA, and Naïve Bayes, with 10-fold cross-validation. Based on the obtained results, SVM outperformed both LDA and Naïve Bayes classification algorithms. Our model will assist clinicians in diagnosing the FTD spectrum and accordingly take reasonable interventions. Although the current accuracy is acceptable, it can be made more robust using a larger dataset. In the future, we can potentially use this approach in the cloud-based automated diagnosis of FTD. Furthermore, we can design a logical explanation generator to validate our proposed method using abductive learning. This generator will identify and provide possible justifications for the correct diagnostic cases, which will improve the accuracy.

APPENDIX A

FreeSurfer [21]

This open-source software FreeSurfer is well-suited for processing and analyzing human brain MRI scans to retrieve volumetric information of different parts of the brain. In these T-weighted images, white matter is lighter; grey matter is darker and cerebrospinal fluid is black. These scans will be used by FreeSurfer to partition the cortical surface and extraction of cortical thickness. Instead of analyzing the brain as a 3D volume, FreeSurfer transforms the cortex into a 2D surface. FreeSurfer has a graphical user interface named Freeview for data analyzing, visualizing, and management. This GUI has much functionality and includes: visualization of FreeSurfer outputs such as the region of interests, volumes, volumetric labeling output, surfaces, cortical thickness, segmentation, and parcellations; also can manually label the MRI scans. We have to perform several preprocessing stages, a few of which are mentioned here.

- **Intensity Normalization**: During image preprocessing, FreeSurfer performs the first step of intensity normalization. This refers to a homogenization of the signal intensity of the grey and white matter to better distinguish between the tissue types and makes it easier to segment the brain. The hypothesis is to manually select the location within the white matter boundary called control points. Voronoi partitioning algorithm is then used to calculate the bias field at non-control points, which allocated the value of the nearest control point. Intensity normalization includes noise correction as well.
- **Skull Stripping:** This is the second step in preprocessing; this removes anything else from the image that is not grey or white matter, such as the eyes, neck, ears, etc. CIVET pipeline is applied for additional correction of skull stripping. FSL is a library used for the analysis and manipulation of MRI brain imaging data. Nipyp**e** provides an interface to use the FSL library of the python module Nibabel. Thus, this is used to skull strip the images in the python code.
- **White Matter Segmentation:** This step segments the white matter from grey matter and cerebrospinal fluid. FreeSurfer includes an algorithm that fixes topological threats in the initial grey and white matter surface derived from a smoothed gradient field. A mesh-based Constrained Laplacian Anatomic Segmentation Using Proximity (CLASP) approach is used for reconstruction of the pial surface.
- **Volumetric Labeling:** By using this step, different regions of the brain are labeled based on volume. It shows volumetric labeling of many subcortical structures such as putamen, Amygdala, hippocampus, ventricles; for attaining correlation among subjects and maintaining the same number of vertices in each hemisphere, the mesh vertices were recomputed.
- **Surface Extraction:** The cortical surface was constructed for both inner and out surface boundaries. When working with surface files, it can be annotation; it will display a collection of labels on a surface, curvature; shows the curvature of the surface, label; will highlight a specific part of the surface.
- **Cortical Thickness:** Cortical thickness was measured between the inner and outer surfaces (Figure 5). Surface-based feature cortical thickness is calculated at each vertex as the mean of the nearest point distance between inner and outer surfaces using linked vertices.

A computational flattening algorithm is used for computation. The mean values of cortical thickness data for neuroanatomically segmented regions manifest as a feature vector.

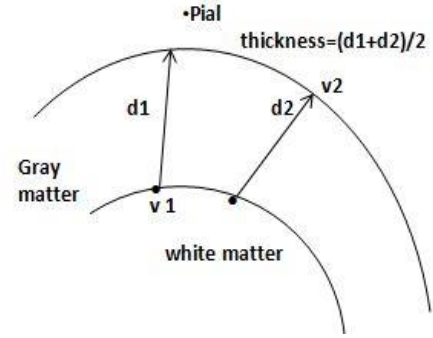

**Fig.5. Distance between two surfaces: the thickness computation described in FreeSurfer**

APPENDIX B

Discriminative Brain Regions

For the classification purpose, we extracted different brain regions, which provided atrophy patterns. We compared two groups based on these extracted brain regions for classification. We obtained the discriminative brain regions as:

$R = \sum_x \times w_1$   (Whereas $\sum_x = XX^T$, X- filtered feature)

R is a multidimensional array which denotes the extracted brain region. Wight vector $w_1$ is defined as the multiplication between dimension reduction component (PCA) matrix and classifier matrix. We are taking surface-based feature cortical thickness in frequency domain, discriminated pattern as:

weight vector $w_1 = Y_{PCA} \times Y_C$

$R_{frequency} = \sum_x \times w_1$